\documentclass{article}
\usepackage{amsmath,amsfonts,amssymb,bm}
\usepackage{graphicx,mlspconf}
\usepackage{algorithmic}
\usepackage{algorithm}
\usepackage{color}
\usepackage{cite}

\title{Visualizing High Dimensional Dynamical Processes}
\name{Andr\'{e}s F. Duque$^1$, Guy Wolf$^2$\sthanks{Work supported by IVADO (l'institut de valorisation des donn\'{e}es)}, Kevin R. Moon$^1$}
\address{$^1$Utah State University, Department of Mathematics \& Statistics, Logan, UT, USA\\
$^2$Universit\'{e} de Montr\'{e}al, Department of Mathematics and Statistics, Montr\'{e}al, QC, Canada
}

\begin{document}
\ninept
%

\maketitle
\begin{abstract}

Manifold learning techniques for dynamical systems and time series have shown their utility for a broad spectrum of applications in recent years. While these methods are effective at learning a low-dimensional representation, they are often insufficient for visualizing the global and local structure of the data. In this paper, we present DIG (Dynamical Information Geometry), a visualization method for multivariate time series data that extracts an information geometry from a diffusion framework. Specifically, we implement a novel group of distances in the context of diffusion operators, which may be useful to reveal  structure in the data that may not be accessible by the commonly used diffusion distances. Finally,  we present a case study  applying our visualization tool to EEG data  to visualize sleep stages.

\end{abstract}
\begin{keywords}
Visualization, dynamical processes, EIG, PHATE, diffusion maps
\end{keywords}
\section{Introduction}
\label{sec:intro}

Manifold learning techniques have become of great interest when studying high dimensional data. The underlying idea behind these methods, is that high dimensional data often encapsulates redundant information. In these cases, the data have an extrinsic dimensionality that is artificially high, while its intrinsic structure is well-modeled as a low-dimensional manifold plus noise. Following the same line of reasoning, dynamical systems and time series can be regarded as processes governed by few underlying parameters, confined in a low-dimensional manifold \cite{lin2006learning, talmon2015manifold}.

In particular, electroencephalographic (EEG) measures can be contemplated in this analytical framework. These measures are taken from different parts of the brain, resulting in a multivariate time series in a high dimensional space. It is known that these time series are highly correlated with each other. Therefore, it can be assumed that there is a low-dimensional representation of the intrinsic dynamics of the brain that can explain a broad spectrum of physical and psychological phenomena such as sleep stages. Additionally, it can be very useful for researchers to achieve meaningful visual representations of this phenomena in two or three dimensions. Such visualizations can be used to better understand the overall shape and finer patterns within the data.

In this paper we present DIG (dynamical information geometry), a dimensionality reduction tool that is designed for visualizing the inherent low-dimensional structure present in high-dimensional dynamical processes. DIG  is built upon a diffusion framework adapted to dynamical processes, followed by an embedding of a novel group of information distances applied to the diffusion operator. The resulting embedding is noise resilient and presents a faithful visualization of the true structure at both local and global scales with respect to time and the overall structure of the data. We demonstrate our DIG on high-dimensional EEG data.

This paper is organized as follows. In Section~\ref{sec:background} we present a brief background of diffusion methods. In Section~\ref{sec:method}, we present the steps of DIG, with a focus on the extension of the diffusion process to dynamical systems and the embedding of information distances. Then, in Section~\ref{sec:experiments}, we show and discuss some practical results for EEG data. And finally in Section~\ref{sec:conclusion}, we conclude our work and propose some extensions.  


\section{Background}
\label{sec:background}
\subsection{Related Work}
Many dimensionality reduction methods exist, some of which have been used for visualization~\cite{maaten2008visualizing,tenenbaum2000isomap,mcinnes2018umap,roweis2000nonlinear,cox2001MDS,moon2000mathematical,van2009dimensionality}. Principal components analysis (PCA)~\cite{moon2000mathematical} and t-distributed stochastic neighborhood embedding (t-SNE)~\cite{maaten2008visualizing} are two of the most commonly used methods for visualization. However, these and other methods are inadequate in many applications. First, these methods tend to favor one aspect of the data at the expense of the other. For example, when used for visualization, PCA typically shows the large scale global structure of the data while neglecting the finer, local structure. In contrast, t-SNE is explicitly designed to focus on the local structure and often distorts the global structure, potentially leading to misinterpretations~\cite{wattenberg2016how}. Second, PCA and t-SNE fail to explicitly denoise the data for visualization. Thus in noisy settings, the true structure of the data can be obscured. In addition, none of these methods are designed to exploit the structure present in dynamical systems.

Diffusion maps (DM) is a popular nonlinear dimensionality reduction technique that effectively denoises the data while capturing both local and global structure~\cite{coifman2006diffusion}. However, DM typically encodes the information in higher dimensions and is thus not optimized for visualization. A more recent visualization method called PHATE was introduced in~\cite{moon2019phate} to exploit the power of DM in denoising and capturing the structure of data while presenting the learned structure in low-dimensions by preserving an information distance between the diffusion probabilities. 

DM has been extended to dynamical systems previously~\cite{lian2015multivariate,talmon2015intrinsic,talmon2013empirical,rodrigues2018multivariate}. In particular, Talmon and Coifman~\cite{talmon2015intrinsic,talmon2013empirical} introduced an approach called empirical intrinsic geometry (EIG) that builds a diffusion geometry using a noise resilient distance. The resulting embedding learned from the geometry is thus noise-free and captures the true structure of the underlying process. However, EIG and other extensions of DM to dynamical systems are still not optimized for visualization as the learned structure of the data is encoded in higher dimensions.  In this work, we introduce a new visualization method DIG that is well-suited for visualizing high-dimensional dynamical processes by preserving an information distance between the diffusion probabilities constructed from a noise resilient distance. This results in a visualization method that represents the true structure of the underlying dynamical process.

EEG signals have been embedded in low dimensional representations for detecting emotional states \cite{wang2014emotional}, pre-seizure states \cite{talmon2015manifold, ataee2007manifold} and sleep dynamics~\cite{rodrigues2018multivariate}. In the latter DM is implemented by building the affinity matrix using both the cross-spectrum distance and the covariance matrix distance as similarity measures between multivariate time series. EIG has also been applied to data including both respiratory and EEG signals~\cite{wu2014assess}. 

\subsection{Preliminaries}
\label{sub:prelim}
Here we provide background on DM~\cite{coifman2006diffusion} and  PHATE~\cite{moon2019phate}. DM learns the geometry of the data by first constructing a graph based on local similarities. The graph is typically constructed by applying a Gaussian kernel to Euclidean distances: 
\begin{align}
    K(\bm{x}_i,\bm{x}_j)=\exp\left( -\frac{||\bm{x}_i-\bm{x}_j||^2}{2\sigma^2}\right), \label{eq:GaussKern}
\end{align}
where $\bm{x}_1,\dots,\bm{x}_N$ are the data and $\sigma$ is a fixed kernel scale or bandwidth that controls the locality scale of the graph. 

The resulting kernel or affinity matrix $K$ encodes the local structure of the data. The process of diffusion is then used to learn the global structure and denoise the data. The first step is to transform the affinity matrix $K$ into a probability transition matrix $P$ (also known as the diffusion operator), by dividing each row by the sum of its entries.  The $i,j$th entry of $P$ indicates the probability of transitioning from $\bm{x}_i$ to $\bm{x}_j$ in a single-step random walk, where the probabilities are calculated based on the relative affinity between points. Hence, if the affinity between two points is high, then the transition probability is high. The global structure of the dynamical process is then learned by performing a $t$-step random walk for integers $t \geq 1$. The transition probabilities of these random walks are given by the $t$-th power of the diffusion operator, (i.e., $P^t$), and rows of this matrix serve as $t$-step diffusion representations of data points. 

In DM, the information encoded in the diffused operator $P^t$ is typically embedded in lower dimensions via eigendecomposition. The Euclidean distances between the embedded coordinates are equivalent to the following scaled distance between entries in the diffused operator:
\begin{align}
    D_t(\bm{x}_i,\bm{x}_j)^2=\sum_{m=1}^N \frac{([P^t]_{mi}-[P^t]_{mj})^2}{\phi_0(m)},
\end{align}
where $\phi_0$ is the stationary distribution of the corresponding Markov chain, or equivalently the first left eigenvector of $P$. 

DM, as described above, has several weaknesses. First, in many applications the data are not sampled uniformly. In these cases, a fixed bandwidth for all points with the Gausssian kernel in (\ref{eq:GaussKern}) may not accurately capture the local data geometry in all settings. For example, a bandwidth tuned for a densely sampled region will be inappropriate for sparsely sampled regions and vice versa.  PHATE counters this by replacing the Gaussian kernel with fixed bandwidth with the $\alpha$-decay kernel with adaptive bandwidth~\cite{moon2019phate}, eq. (\ref{eq:AdaGaussKern}). The adaptive bandwidth enables the affinities to scale with the local density while the $\alpha$-decay kernel corrects inaccuracies that may be introduced by the adaptive bandwidth in sparse sampled regions:

{\scriptsize
\begin{align}
    K_{k,\alpha}(\bm{x}_i,\bm{x}_j)=\frac{1}{2}\exp\left( -\frac{||\bm{x}_i-\bm{x}_j||^2}{\sigma_{k}(\bm{x}_i)}\right)^\alpha + \frac{1}{2}\exp\left( -\frac{||\bm{x}_i-\bm{x}_j||^2}{\sigma_{k}(\bm{x}_j)}\right
    )^\alpha. \label{eq:AdaGaussKern}
\end{align}}

\noindent{}The value of $\sigma_{k}(\bm{x}_i)$ is the distance from $\bm{x}_i$ to its $k$-nearest neighbor. Since this value changes for different observations, it may result in a non-symmetric kernel. Taking the average, as shown in eq. (\ref{eq:AdaGaussKern}), mitigates this issue. 

Second, choosing an appropriate time scale $t$ is a difficult problem, as this parameter controls the resolution of the captured diffusion representation. In PHATE, the von Neumann entropy (VNE) is used to automatically tune this scale. The VNE (also known as spectral entropy) for each $t$ is the entropy of eigenvalues of the diffused operator $P^t$ (normalized to form a probability distribution), and provides a soft proxy for the number of significant eigenvalues of $P^t$. As $t\rightarrow\infty$, the VNE converges to zero, since the diffusion process is constructed to have a unique stationary distribution. The rate of decay of the VNE as $t$ increases is thus used to determine the appropriate value of $t$ at the transition between rapid decay, which is interpreted in~\cite{moon2019phate} as corresponding to the elimination of noise, and slow decay (interpreted there as losing meaningful information).

Third, DM is not well-suited for visualization as the eigendecomposition does not explicitly force as much information as possible into low dimensions. In fact, when the data have a branching structure, DM tends to place different branches in different dimensions~\cite{haghverdi2016diffusion}. Additionally, attempts to directly embed the diffusion distances into low dimensions for visualization using multidimensional scaling (MDS)~\cite{cox2001MDS} can lead to unstable and inaccurate embeddings~\cite{moon2019phate}. PHATE counters this by constructing a potential distance (i.e., comparing energy potentials) from diffused probabilities and then directly embedding potential distances in two or three dimensions using MDS. 





\section{The DIG Algorithm}
\label{sec:method}



In this section, we extend principles of DM and PHATE to dynamical systems to derive DIG. In this context, we present a family of information distances and derive some of their properties. 

\subsection{Diffusion with Dynamical Systems}
In the context of dynamical systems it is needed to learn the local structure by constructing a matrix that encodes the local distances between data points. These local distances can be taken as an input for (\ref{eq:AdaGaussKern})  to build a diffusion operator, from which information is extracted for visualization. To do this, we build upon the EIG framework~\cite{talmon2013empirical,talmon2015intrinsic} which uses a state-space formalism (\ref{eq1})-(\ref{eq2}):

\begin{align}
    \bm{z}_{t} &=  \bm{y}_{t}(\bm{\theta}_{t}) + \bm{\xi}_{t} \label{eq1} \\
    d\theta^{i}_{t} &= a^{i}(\theta^{i}_{t})dt + dw^{i}_{t}, \ \ i=1,\ldots,d. \label{eq2}
\end{align}

The multivariate time series $\bm{z}_t$ represents the observed time series data while $\bm{\theta}_t$ represents the hidden (unobserved) states that drive the process. $\bm{z}_t$ can be viewed as a corrupted version of a clean process $\bm{y}_t$ that is driven by the hidden states, where the corruption $\bm{\xi}_t$ is a stationary process independent of $\bm{y}_t$.
In general, we can view $\bm{y}_t$ as being drawn from a conditional pdf $p(\bm{y}|\bm{\theta})$. In the stochastic process (\ref{eq2}), the unknown drift functions $a^{i}$ are independent from $\theta^{j}$, $j \neq i$. Therefore, we assume local independence between $\theta^{i}_{t}$ and $\theta^{j}_{t}$, $\forall i \neq j$. The variables $w^{i}_{t}$ are Brownian motions.

It can be shown that the pdf $p(\bm{z}|\bm{\theta})$ is a linear transformation of $p(\bm{y}|\bm{\theta})$~\cite{talmon2013empirical,talmon2015intrinsic}. Since the pdfs are unknown, we use histograms as their estimators. Each histogram $\bm{h_{t}} = (h_{t}^{1}, \ldots, h_{t}^{Nb})$ has $Nb$ bins, and is built with the observations within a time window of length $L_{1}$, centered at $\bm{z_{t}}$. The expected value of the histograms, e.g.\ $\mathbb{E}(h_{t}^{j})$, is a linear transformation of $p(\bm{z}|\bm{\theta})$. Since the Mahalanobis distance is invariant under linear transformations, it can be deduced that the distance (\ref{mah_dis}) is noise resilient~\cite{talmon2015intrinsic}:

{\footnotesize
\begin{align}
    d^{2}(\bm{z}_{t}, \bm{z}_{s}) = (\mathbb{E}(\bm{h}_{t}) - \mathbb{E}(\bm{h}_{s}))^{T}(\bm{C}_{t} + \bm{C}_{s})^{-1} (\mathbb{E}(\bm{h}_{t}) - \mathbb{E}(\bm{h}_{s})),
    \label{mah_dis}
\end{align}}%
where $C_{t}$ and $C_{s}$ are the covariance matrices in the histograms space, in a time window of length $L_{2}$, centered at $\bm{h}_{t}$ and $\bm{h}_{s}$, respectively. 
Also, it can be proved under certain assumptions that $d(\bm{z}_{t}, \bm{z}_{s})$ is a good approximation of the distance between the underlying state variables~\cite{talmon2015intrinsic}: 

\begin{align}
    \lVert \bm{\theta}_{t} - \bm{\theta}_{s} \rVert^2 \approx d^{2}(\bm{z}_{t}, \bm{z}_{s}).
    \label{dis_state}
\end{align}

To learn the global relationships from the local information encoded by the distances in eq. (\ref{mah_dis}), we use the diffusion process. We apply the $\alpha$-decay kernel in eq. (\ref{eq:AdaGaussKern}) to the Mahalanobis distances in eq. (\ref{mah_dis}), where the $k$-nearest neighbor distances are also calculated with this distance. We use the $\alpha$-decay kernel to account for potentially different regions of density, as described in Section~\ref{sub:prelim}. The diffusion operator $P$ is then constructed by row-normalizing the resulting kernel matrix as before.

For comparison purposes, we also make use of an alternative distance to (\ref{mah_dis}).  Assuming that the data within time windows of length $L_{1}$ centered at $z_{t}$ follows a multivariate Gaussian distribution $\mathcal{N}(\bm{\mu}, \Sigma_{t})$, we can compute the geodesic distance between different time windows of data using the Fisher information as the Riemannian metric \cite{atkinson1981rao} as follows: 

\begin{align}
     d^{2}(\bm{z}_{t}, \bm{z}_{s}) = \frac{1}{2}\sum_{i=1}^{N}\text{ln}(\lambda_{i}),
     \label{eq:geodesicGaussian}
\end{align}
where $\lambda_{i}$ are the roots of, $|\Sigma_{t} - \lambda\Sigma_{s}| = 0$. The diffusion operator can then be obtained using this distance as input to the $\alpha$-decay kernel.



\subsection{Embedding Information Distances}
Typically, information is extracted from the diffusion operator $P$ by either eigendecomposition or by embedding the diffusion distances. However, as mentioned previously, the former typically fails to provide a low-dimensional representation that is sufficient for visualization while the latter can result in unstable embeddings in some cases~\cite{moon2019phate}. To overcome this, DIG extracts the information from the diffusion operator by embedding an information distance instead. We focus on a broad family of information distances that are parametrized by $\gamma$: 
\begin{align}
    {\scriptstyle D^{\gamma}_t(\bm{z}_i,\bm{z}_j)^2}=
     \left\{
\begin{array}{ll}
      \sum_{m=1}^N \frac{(\text{log}[P^t]_{mi}-\text{log}[P^t]_{mj})^2}{\phi_0(m)}, &  \scriptstyle{\gamma =1}\\
      \sum_{m=1}^N \frac{([P^t]_{mi}-[P^t]_{mj})^2}{\phi_0(m)}, &  \scriptstyle{\gamma = -1} \\
      \sum_{m=1}^N \frac{2(([P^t]_{mi})^{\frac{1-\gamma}{2}}-([P^t]_{mj})^{\frac{1-\gamma}{2}})^2}{(1-\gamma)\phi_0(m)}, &  \scriptstyle{-1 < \gamma < 1}. \\
\end{array}  \right. 
\label{infdis}
\end{align}
The parameter $\gamma$ controls the level of influence of the lower differences among probabilities in the overall distance. For example, the standard diffusion distances ($\gamma = -1$) are highly influenced by the highest absolute differences among probabilities. In contrast, the potential distances ($\gamma = 1$), which were used in PHATE, account for the relative differences between them. Thus the standard diffusion distances and the potential distances can be viewed as two extremes of a general class of distances over the diffusion geometry.

\begin{algorithm}[H]
\begin{algorithmic}[1]

\renewcommand{\algorithmicrequire}{\textbf{Input:}} \renewcommand{\algorithmicensure}{\textbf{Output:}}

\REQUIRE Data matrix $X$, neighborhood size $k$, locality scale $\alpha$, time windows length $L_{1}$ and $L_{2}$, number of bins $Nb$, information parameter $\gamma$, desired embedding dimension $m$ (usually 2 or 3 for visualization)

\ENSURE The DIG embedding $Y_m$

\STATE $D\leftarrow$ compute pairwise distance matrix from $X$ using distance (\ref{mah_dis})

\STATE $K_{k,\alpha}\leftarrow$ compute local affinity matrix from $D$ and $\sigma_{k}$

\STATE $P\leftarrow$ normalize $K_{k,\alpha}$ to form a Markov transition matrix (diffusion operator)

\STATE $t\leftarrow$ compute time scale via Von Neumann Entropy~\cite{moon2019phate} 

\STATE Diffuse $P$ for $t$ time steps to obtain $P^t$


\STATE $\mathfrak{V}^t\leftarrow$ compute the information distance matrix in eq.~\ref{infdis} from $P^{t}$ for the given $\gamma$

\STATE $Y'\leftarrow$ apply classical MDS to $\mathfrak{V}^t$

\STATE $Y_{m}\leftarrow$ apply metric MDS to $\mathfrak{V}^t$ with $Y'$ as an initialization

\end{algorithmic}

\caption{The DIG algorithm}

\label{alg:dig}
\end{algorithm}





It can be shown that for $\gamma\in[-1,1]$, the distance $D_t^\gamma$ forms an M-divergence~\cite{salicru1985ciertas,salicru1995entropy}. Furthermore, when $\gamma=0$, $D_t^\gamma$ becomes proportional to the Hellinger distance, which is an $f$-divergence~\cite{csiszar1964informationstheoretische,ali1966general}. Further, $f$-divergences are directly related to the Fisher information and thus are well-suited for building an information geometry~\cite{berisha2014empirical}. Therefore, $f$-divergences may be desirable for embedding the diffused probabilities.


We also consider another information distance based on $f$-divergences that has not been applied to diffusion operators as far as we know. Since the rows of the diffusion matrix $P$ can be interpreted as multinomial distributions, we can compute the geodesic distance between them using the Fisher information as the Reimmanian metric \cite{amari2016information}. This can be seen as an extension of the KL divergence or the Hellinger distance, both $f$-divergences, for distributions far apart from each other.  Such distances are as follows  \cite{lafferty2005diffusion, atkinson1981rao}: 


\begin{align}
    D(\bm{z}_i,\bm{z}_j) = 2\text{cos}^{-1}\left(\sum_{m=1}^{N}\sqrt{[P^t]_{mi}[P^t]_{mj}}\right).
    \label{infgeo}
\end{align}

After the information distances have been obtained, DIG applies metric MDS to the information distances to obtain a low-dimensional representation. Given an information distance $D$, metric MDS minimizes the following stress function:
\begin{equation}
\label{eq:stress}
\mathtt{Stress}(\hat{z}_1,\ldots,\hat{z}_N)=\sqrt{\frac{\sum_{i,j}\left(D(\bm{z}_i,\bm{z}_j)-\|\hat{x}_i-\hat{x}_j\|\right)^2 }{ \sum_{i,j}\left(D(\bm{z}_i,\bm{z}_j)\right)^2}},
\end{equation}
where the $\hat{z}_i$ are the $m$-dimensional embedded coordinates. For visualization, $m$ is chosen to be 2 or 3. See Algorithm \ref{alg:dig} for pseudocode summarizing the described steps of DIG.



\section{Experimental results}
\label{sec:experiments}

\begin{figure*}[htb]
  \includegraphics[width=\textwidth]{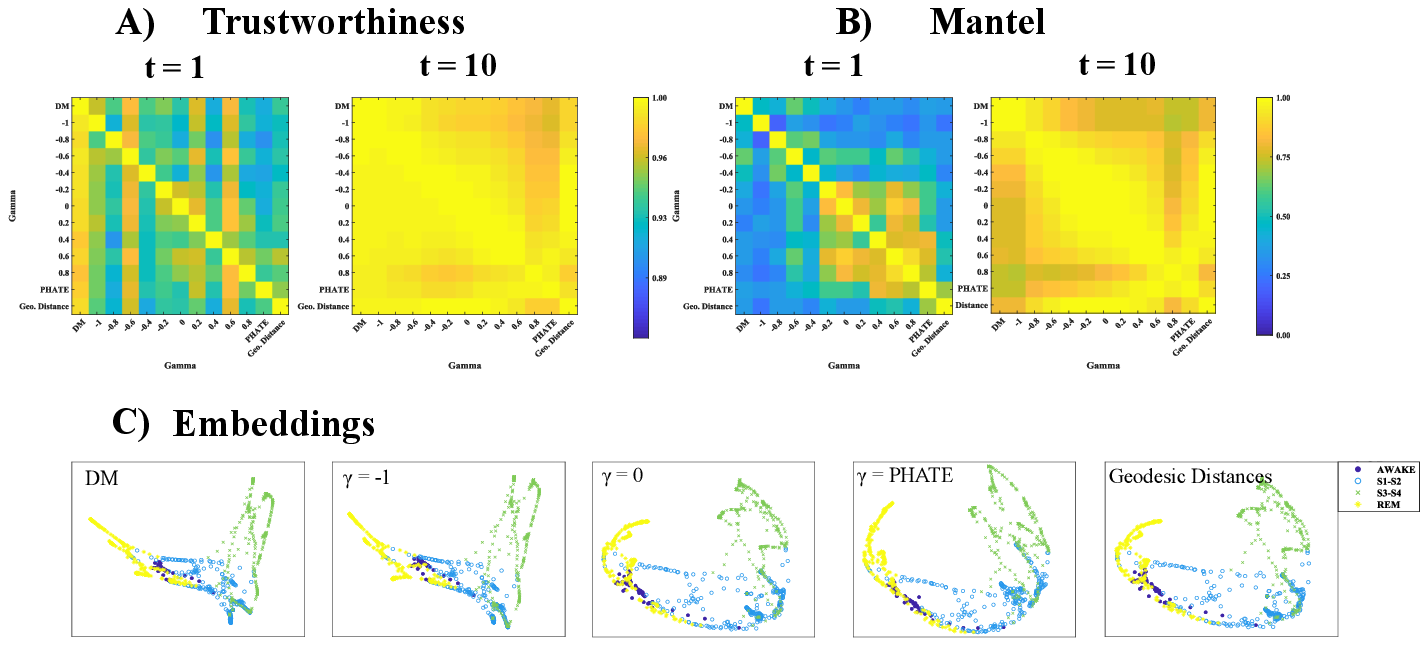}
   \caption{\footnotesize Impact of $\gamma$  on the visualizations of EEG data from~\cite{terzano2002atlas,goldberger2000physiobank} using the Mahalanobis distance (\ref{mah_dis}). \textbf{(A)} The relative local distortion, measured by the Trustworthiness between embeddings using different values of $\gamma$, as well as the embeddings generated by the usual eigendecomposition (DM) and distance eq. (\ref{infgeo}) (Geodesic distances). \textbf{(B)} The relative global distortion, measured by the Mantel test correlation coefficient between different embeddings as in \textbf{(A)}. \textbf{(C)} 2-dimensional embeddings for different values of $\gamma$, colored by sleep stages. The DM embedding is obtained by the eigendecomposition of $P^{t}$, while the $\gamma = -1$ uses diffusion distances embedded with MDS. The similarity among the embeddings is visually apparent in \textbf{(C)}, as well as quantitatively from the near to 1 values across the heatmaps for the Trustworthiness and Mantel measures when $t=10$.} 
  \label{Tw_emb}
\end{figure*}

\begin{figure*}[ht]
  \includegraphics[width=\textwidth]{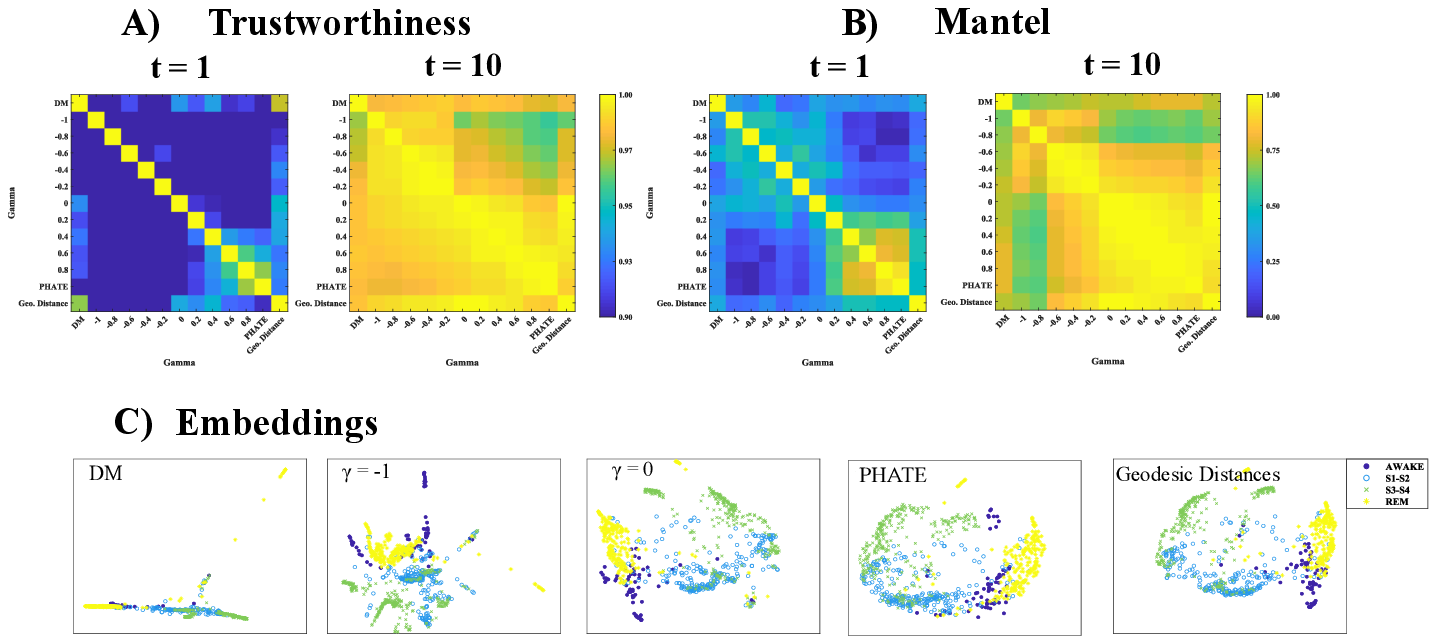}
    \caption{ \footnotesize Impact of $\gamma$ on the visualizations of EEG data from~\cite{terzano2002atlas,goldberger2000physiobank} using the Gaussian-based geodesic distance (\ref{eq:geodesicGaussian}). \textbf{(A)} The relative local distortion, measured by the Trustworthiness between embeddings using different values of $\gamma$, as well as the embeddings generated by the usual eigendecomposition (DM) and distance eq. (\ref{infgeo}) (Geodesic distances). \textbf{(B)} The relative global distortion, measured by the Mantel test correlation coefficient between different embeddings as in \textbf{(A)}. \textbf{(C)} 2-dimensional embeddings for different values of $\gamma$, colored by sleep stages.  The differences between DM, $\gamma = -1$ and $\gamma = -0.8$, with the others, become apparent in light of these three subplots.}
  \label{emb2}
\end{figure*}

\subsection{Data}

We now present a real-world data application using EEG data provided by \cite{terzano2002atlas,goldberger2000physiobank}. The original data is sampled at 512Hz and labeled for every 30 second interval, within six sleep categories according to R\&K rules (REM, Awake, S-1, S-2, S-3, S-4). Due to the lack of observations in some stages,  we group S-1 with S-2, and S-3 with S-4. We band-filtered the data between 8-40 Hz, and down-sampled it to 128Hz.  

\subsection{Experimental Setup}

The tuning and influence of the parameters $\alpha$, $t$ and $k$ for nondynamical data have been covered in \cite{moon2019phate}.  For the EEG data, we found that the visualizations are  highly robust for a wide range of configurations. Preliminary experiments showed that $k=5$, $\alpha = 10$, and $t = 10$ give meaningful results. Therefore, we used these parameters for all experiments shown here unless otherwise stated. To compute the distances (\ref{mah_dis}) and (\ref{eq:geodesicGaussian}), we need to choose $L_{1}$, the window size. Its selection is driven by the way the data is presented. In our case, we simply took $L_{1} = 3840$, the number of observations in the 30s span. We also selected $L_{2} = 10$, and the number of bins for the histograms $Nb = 20$. We do not focus on these parameters since  their impact have been already studied in the cited literature.

Conversely, as far as our knowledge goes, there has not been any previous study addressing the properties of the information distances mentioned in the previous section. Thus, we focus on how these distances may affect the learning process.

For this purpose, we wish to measure the relative distortion of the embeddings both locally and globally, when setting different values of $\gamma$, as well as using the geodesic distance in eq. (\ref{infgeo}). One commonly used approach to determine the local distortion, is the Trustworthiness proposed in \cite{kaski2003trustworthiness}. This measure provides a penalty  when one of the $k$-nearest neighbors of an observation in the low-dimensional embedding is not one of the $k$-nearest neighbors in the original data space. For our particular case, we compare the low-dimensional embeddings with each other. Trustworthiness gives an index that goes from zero to one. The lower the distortion is, the closer Trustworthiness gets to one. Notice the trustworthiness does not need to be symmetric.

To measure global differences between embeddings, we employed the  Mantel test \cite{mantel1967detection}, which gives a level of similarity between the distance matrices in the embedded dimensions. This gives an overall measure of the similarity between embeddings. The motivation for this test is the fact that distances are not independent from each other. For example, changing the position of a single observation will result in the distortion of $N-1$ distances. This implies that the direct calculation of the correlation between distances may not be enough  to accurately assess the similarity between distance matrices. The Mantel test takes such dependencies between distances into account. The test outputs a correlation coefficient between 0 and 1, that can be interpreted similarly as the usual Pearson correlation. 

\subsection{Discussion}

Figure \ref{Tw_emb}\textbf{AB} shows how using distance (\ref{mah_dis}), the embedding becomes robust with respect to $\gamma$ as $t$ increases. From a local perspective, the Trustworthiness measure among embeddings with different values of $\gamma$ gets closer to one using $t=10$ than for $t=1$. This shows a high local similarity independent of the value of gamma. From a global perspective, the Mantel test also shows a high dissimilarity for $t=1$, while for $t=10$ the embeddings become more similar. The previous assessments can be visually corroborated by looking at the embeddings themselves in Figure~\ref{Tw_emb}\textbf{C}. Finally, in Figure~\ref{Tw_emb}\textbf{C} we can observe a good representation of the sleep dynamics as there is a visually clear discrimination among sleep stages, especially for higher values of $\gamma$ or when using the diffusion geodesic distances. For these higher values of $\gamma$, the central structure of the embedding is more clearly defined than when using DM or lower values of $\gamma$. But largely, the embedding is robust to the choice of $\gamma$, suggesting that the specific choice of information distance is not too important when using the Mahalanobis distance in eq.~(\ref{mah_dis}) from the EIG framework. Furthermore, the results suggest that the use of the selected $f$-divergences does not provide any apparent advantage in this setting over the other information distances.

For the previous case, different values of $\gamma$ produced largely the same results. This is no longer the case when we use the Gaussian-based geodesic distance in eq.~\ref{eq:geodesicGaussian} as shown in Figure~\ref{emb2}. First, Figure~\ref{emb2}\textbf{A} and Figure~\ref{emb2}\textbf{B} show quantitatively a greater difference between the embeddings than in Figure \ref{Tw_emb}. This can be visually confirmed by looking at the embeddings in  Figure~\ref{emb2}\textbf{C}. In this case, the traditional DM tends to condense the structure together, and the use of the alternative $\gamma$ values may reveal more details of the structure of the data. The most left embedding is a clear representation of such a situation, where DM does not show a suitable discrimination of the sleep stages. But when increasing the value of gamma, a more suitable representation is achieved.     

\begin{figure}
    \centering
    \includegraphics{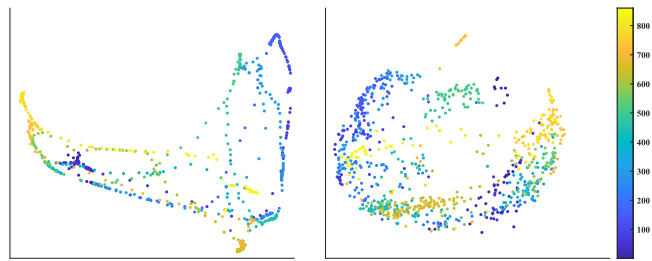}
    \caption{\footnotesize  Visualization of the EEG data from~\cite{terzano2002atlas,goldberger2000physiobank} colored by time steps using distance (\ref{mah_dis}) at the left, and distance (\ref{eq:geodesicGaussian}) at the right. Here we see how the left visualization presents a more denoised version, with  clearer time-evolving transitions.}
    \label{fig:time_steps}
\end{figure}

There are clear visual differences between the embeddings implementing distances (\ref{mah_dis})  and (\ref{eq:geodesicGaussian}). In Figure~\ref{fig:time_steps}, we show a comparison of two embeddings colored by time steps. The left embedding is built using distance (\ref{mah_dis}),  achieving a good, clean visualization of the  process across time. In fact, the different branches in the central region of the embedding are created from different periods of time, suggesting that transitions from different sleep stages may differ slightly depending on the total sleep time. In contrast, the Gaussian-based geodesic distance (\ref{eq:geodesicGaussian}), displayed in the right embedding, not only inherits more of the original noise but also obscures the path of the process across time. Thus, using the Mahalanobis distance appears to better denoise the data and preserve the overall structure and time progression of this data.

\section{Conclusion}
\label{sec:conclusion}
In this work, we introduced a manifold learning tool for visualizing dynamical processes based on a diffusion framework. We addressed some of the shortcomings of the traditional diffusion maps approach for visualization, and used elements from PHATE and EIG to overcome them. We showed that when using the EIG-based distance, the visualization is robust to the choice of information distance. 

We presented experimental results where we were able to discover sleep dynamics using solely EEG recordings. The 2-dimensional visualizations showed a clear distinction among sleep stages, as well as the time-varying progress of the processes. 

Future work  includes extending the analysis and comparison between different distance measures for time series data, further analyzing the impact of different information distances, and applying DIG to financial and biological data.



\bibliographystyle{IEEEbib}
\bibliography{VHDDP}

\end{document}